\documentclass[conference]{IEEEtran}
\usepackage{hyperref}	
\usepackage{setspace}

\IEEEoverridecommandlockouts
\usepackage{amsmath,amssymb,amsfonts}
\usepackage{algorithmic}
\usepackage{graphicx}
\graphicspath{ {./Figures/} }
\usepackage[numbers]{natbib}

\usepackage{textcomp}
\usepackage{xcolor}
\usepackage{booktabs}
\usepackage{floatrow}
\usepackage{array, multirow}
\usepackage{makecell}
\usepackage{tabularx}
\usepackage{dblfloatfix}
\usepackage{lipsum}

\def\BibTeX{{\rm B\kern-.05em{\sc i\kern-.025em b}\kern-.08em
    T\kern-.1667em\lower.7ex\hbox{E}\kern-.125emX}}

\makeatletter
\def\@IEEEpubidpullup{4\baselineskip}
\makeatother
\begin{document}

\IEEEpubidadjcol
\IEEEoverridecommandlockouts
\IEEEpubid{
	\parbox{\columnwidth}{\vspace{-4\baselineskip}Permission to make digital or hard copies of all or
part of this work for personal or classroom use is granted without fee provided that copies are not
made or distributed for profit or commercial advantage and that copies bear this notice and the full
citation on the first page. Copyrights for components of this work owned by others than ACM must
be honored. Abstracting with credit is permitted. To copy otherwise, or republish, to post on servers
or to redistribute to lists, requires prior specific permission and/or a fee. Request permissions from
\href{mailto:permissions@acm.org}{permissions@acm.org}.\hfill\vspace{-0.8\baselineskip}\\
		\begin{spacing}{1.2}
			\small\textit{ASONAM '19}, August 27--30, 2019,	Vancouver, BC, Canada	\\
			\copyright\space2019 Association	for	Computing	Machinery.	\\
			ACM	ISBN	978-1-4503-6868-1/19/08...\$15.00	\\
			\url{http://dx.doi.org/10.1145/3341161.3342892}	
		\end{spacing}
		\hfill}
	\hspace{0.9\columnsep}\makebox[\columnwidth]{\hfill}}
\IEEEpubidadjcol

\title{Neural Language Model Based Training Data Augmentation for Weakly Supervised Early Rumor Detection}

\author{\IEEEauthorblockN{Sooji Han}
\IEEEauthorblockA{\textit{Department of Computer Science} \\
\textit{The University of Sheffield}\\
Sheffield, UK \\
sooji.han@sheffield.ac.uk}
\and
\IEEEauthorblockN{Jie Gao}
\IEEEauthorblockA{\textit{Department of Computer Science} \\
\textit{The University of Sheffield}\\
Sheffield, UK \\
j.gao@sheffield.ac.uk}
\and
\IEEEauthorblockN{Fabio Ciravegna}
\IEEEauthorblockA{\textit{Department of Computer Science} \\
\textit{The University of Sheffield}\\
Sheffield, UK \\
f.ciravegna@sheffield.ac.uk }}

\maketitle

\begin{abstract}
The scarcity and class imbalance of training data are known issues in current rumor detection tasks. We propose a straight-forward and general-purpose data augmentation technique which is beneficial to early rumor detection relying on event propagation patterns. The key idea is to exploit massive unlabeled event data sets on social media to augment limited labeled rumor source tweets. This work is based on rumor spreading patterns revealed by recent rumor studies and semantic relatedness between labeled and unlabeled data. A state-of-the-art neural language model (NLM) and large credibility-focused Twitter corpora are employed to learn context-sensitive representations of rumor tweets. Six different real-world events based on three publicly available rumor datasets are employed in our experiments to provide a comparative evaluation of the effectiveness of the method. The results show that our method can expand the size of an existing rumor data set nearly by 200\% and corresponding social context (i.e., conversational threads) by 100\% with reasonable quality. Preliminary experiments with a state-of-the-art deep learning-based rumor detection model show that augmented data can alleviate over-fitting and class imbalance caused by limited train data and can help to train complex neural networks (NNs). With augmented data, the performance of rumor detection can be improved by 12.1\% in terms of F-score. Our experiments also indicate that augmented training data can help to generalize rumor detection models on unseen rumors.  

\end{abstract}


\section{Introduction}\label{intro}

Research areas that have recently been received much attention in using Machine Learning (ML) and Natural Language Processing for automated rumor and fake news detection ~\citep{Kwon17Rumor,Wong2018Rumor} and fact-checking ~\citep{Boididou2018Verifying,kochkina2018all}. In particular, deep learning architectures have been increasingly popular by providing significant improvements to state-of-the-art (SoA) performances. Despite their success, several challenges have yet to be tackled. One major bottleneck of state-of-the-art ML methods for rumor studies is that they require a vast amount of labeled data samples to be trained. However, the manual annotation of large-scale and noisy social media data for rumors is highly labor-intensive and time-consuming \citep{Zubiaga2016LearningRD} as it requires deeper domain knowledge and a more elaborate examination than common annotations like image tagging or named entity annotations do. Due to limited labeled training data, existing NNs for rumor detection usually have shallow architecture \citep{chen2018call,ma2016detecting}. This restricts a further exploration of NNs for representation learning through many layers of nonlinear processing units and different levels of abstraction \citep{zhong2016overview}, which results in over-fitting and generalization concerns. The scarcity of labeled data is a major challenge facing for the research of rumors in social media \citep{aker2017stance}.  Another problem is that publicly available data sets for rumor-related tasks suffer from imbalanced class distributions \citep{liu2017rumors,kochkina2018all}. Existing methods for handling the class imbalance problem (e.g., oversampling and the use of synthetic data \citep{xu2015scalable}) may cause over-fitting and poor generalization performance. A methodology for larger rumor training set with the minimum of human supervision is necessary.

Data augmentation is the key to learning with modern deep neural networks (DNNs) as they require a large amount of data for training. The artificial augmentation of training data helps to alleviate data sparseness and class imbalance, reduce over-fitting, and reduce generalization error, thereby sustaining deeper networks and improving their performance. We argue that enriching existing labeled rumor data with duplicated (but unique) tweets or corresponding variants is a promising attempt for early rumor detection methods \citep{zubiaga2017exploiting} that rely on the structure of rumor propagation. Recent findings \citep{maddock2015characterizing, chen2018call} show that rumors spread via the distribution of original sources. Original sources can quickly evolve into several new variants within the first few minutes in social media. Variations will gradually be increased with more information such as URLs (links) and photos by users. Links are usually created as new messages without attribution. Although new variations of rumors do not usually have any link or acknowledgment of their original sources, they can increase the credibility of sources with low credibility and the likelihood of rumor spreading. Malicious users leverage users' trust to spread rumors and harmful content on social media \citep{baxter2015social,arif2016information}. According to previous studies on rumors on social media \citep{maddock2015characterizing,enquiring15zhao}, new variations of rumors posted within the first few peaks in event diffusion are mostly textual variants. 
80\% of a publicly available rumor tweet corpus consists of duplicated contents on average \citep{chen2018call}. Previous studies revealed that variations of rumors share similar propagation patterns, and proposed methods for identifying rumors based on temporal, structural, and linguistic properties of their propagation \citep{liu2017rumors, Kwon17Rumor}.

In this paper, we propose a novel data augmentation method for automatic rumor detection based on semantic relatedness. The method is based on a publicly available paraphrase identification corpus, context-sensitive embeddings of labeled reference tweets and unlabeled candidate source tweets. Pairwise similarity is used to guide the assignment of pseudo-labels to unlabeled tweets. ELMo \citep{Peters2018deep}, a state-of-the-art context-sensitive NLM, is fine-tuned on a large credibility-focused social media corpus and used to encode tweets. Our results show that data augmentation can contribute to rumor detection via deep learning with increased training data size and a reasonable level of quality. This has potential for further performance improvements using deeper NNs. We present data augmentation results for six real-world events and the performance of a state-of-the-art DNN model for rumor detection with augmented data in Section \ref{results}.  The augmented rumor corpus is available via \url{https://zenodo.org/record/3269768}

\section{Related Work}\label{relatedwork}

Automatic data augmentation has been employed in a wide range of ML tasks as it helps to improve the generalization performance of deep learning models. Data augmentation usually makes use of transformations to the training set. For example, common transformations for images include flipping, rotating, scaling, cropping, and adding noises. Our work focuses on data augmentation for textual data. The most common approach for augmenting textual data is to replace words or phrases with synonyms. In one work on text classification~\cite{zhang2015character}, a WordNet thesaurus, in which synonyms for a word or phrase are grouped and ordered by semantic relatedness, is used to replace words in training corpora including reviews, news articles, and DBpedia data sets. The number of words to be replaced and an integer position in the index of synonyms of a given word are randomly determined from a geometric distribution with parameter $p=0.5$. The authors present that augmented data improves the performance of convolutional neural networks (CNNs) for text classification. In particular, character-level CNNs trained on augmented data achieves the best performance. Recent research \cite{vosoughi16tweet, vijayaraghavan2016deepstance} applies this method to tweets, and shows that data augmentation can bring performance gains in deep learning tasks on noisy and short social media texts. Vosoughi et al.\cite{vosoughi16tweet} augment domain-independent English tweets for training an encoder-decoder embedding model built with character-level CNN and long short-term memory (LSTM). The number of tweets before data augmentation is not presented, but the author report that 3 million tweets in total are available after data augmentation. Another work \cite{vijayaraghavan2016deepstance} on tweet stance classification employs the same technique but uses Word2Vec instead of the WordNet thesaurus to replace words in text. Synonyms of a given word are ranked based on cosine similarity between the Word2Vec vector of given word and that of each synonym. The reported number of augmented tweets is 500,000. Despite a wide use of synonyms in text data augmentation and their contribution to performance enhancement, the use of paradigmatic relations can provide a wider range of substitutes for a given word \cite{kobayashi2018contextual}. To this end, Kobayashi \cite{kobayashi2018contextual} proposes methods for context-aware data augmentation based on a conditional bi-directional language model (BiLM). BiLM computes the probability distribution of possible substitutes for a given word in a sentence based on its context (i.e., a sequence of surrounding words). Their method is evaluated for text classification using six different data sets including movie reviews and answer types of questions. Contextual data augmentation makes marginal improvements over performances of synonym-based methods. Recently, a data augmentation method which combines $n-$grams and Latent Dirichlet Allocation (LDA) has been proposed \cite{abulaish19text}. The method is evaluated on its effectiveness in polarity classification (negative or positive) of reviews using CNNs. LDA is used to extract and rank keywords from positive and negative review corpora separately. Variations of a review are created by combining the original review with its trigrams that contain at least one keyword from the LDA review keywords of the same class type (i.e., positive or negative). Whereas most work on text data augmentation generates variations of a text based on the transformation of words and phrases, a recent work augments tweets by translating a tweet to a different language and then translating it back to the original language \cite{luque2018atalaya}. Unlike current artificial data augmentation methods based on modifications to existing data or reliance on limited knowledge bases, our method uses large-scale real-world social media data. It can not only increase the amount of training data, but most importantly help to increase the quality and diversity of original data.

\section{Data} \label{data}
We use three publicly available rumor datasets covering a wide range of real-world events on social media, a Twitter paraphrase corpus, and two large-scale Twitter corpora.\\
\noindent\textbf{PHEME(6392078)~\citep{kochkina2018all}} This data consists of manually labeled rumors and non-rumors for 9 events. It is used as a reference data for data augmentation (see details in Section~\ref{overview}).\\
\textbf{CrisisLexT26~\citep{Olteanu2015What}} This data comprises tweets associated with 26 hazardous events happened between 2012 and 2013. A subset of data is manually labeled based on informativeness, information types, and information sources. This data is used as a reference data for data augmentation (see Section~\ref{overview})\\ 
\textbf{Twitter event datasets (2012-2016)~\citep{Zubiaga2018Longitudinal}} 
This data consists of over 147 million tweets associated with 30 real-world events unfolded between February 2012 and May 2016. We use this data as a pool of candidate source tweets. We choose six out of 30 available events, for which we can generate references corresponding to the candidate pool including ``Ferguson unrest'', ``Sydney siege', ``Ottawa shootng'', ``Charliehebdo'', ``Germanwings crash'', and ``Boston marathon bombings''. We refer to five events except the `Boston marathon bombings' as \textbf{`PHEME5'}. (see Section~\ref{augref})

\noindent \textbf{SemEval-2015 task 1 data~\citep{Xu2014Extracting}} This data is built for paraphrase identification and semantic similarity measurement. It is employed in our semantic relatedness method to fine-tune an optimum relatedness threshold through a pairwise comparison between the embeddings of labeled reference tweets and those of unlabeled candidates event tweets (see Section~\ref{thresh-finetune}).\\
\noindent \textbf{CREDBANK~\citep{mitra2015credbank}} This data comprises more than 80M tweets grouped into 1049 real-world events, each of which were manually annotated with credibility ratings. This large corpus is leveraged to fine-tune ELMo model in order to provide better representations for rumor-related tasks (refer to Section~\ref{method-data}).\\
\textbf{SNAP data~\citep{yang2011patterns}} The SNAP Standford Twitter data ``twitter7'' \footnote{We downloaded the dataset from https://snap.stanford.edu/data/twitter7.html (accessed on March, 2019)} is used as a general purpose Twitter corpus in our experiment. This is a collection of 476 million tweets collected between June-Dec 2009. We use this dataset to conduct comparative analysis of effectiveness of \textit{CREDBANK} as a rumor task specific dataset for language model training. See Section ~\ref{method-data} for the details of a post-processed corpus.

\section{Methodology}\label{methods}
\subsection{Overview of the proposed method}\label{overview}

\begin{figure*}[h!]
  \includegraphics[width=0.7\textwidth]{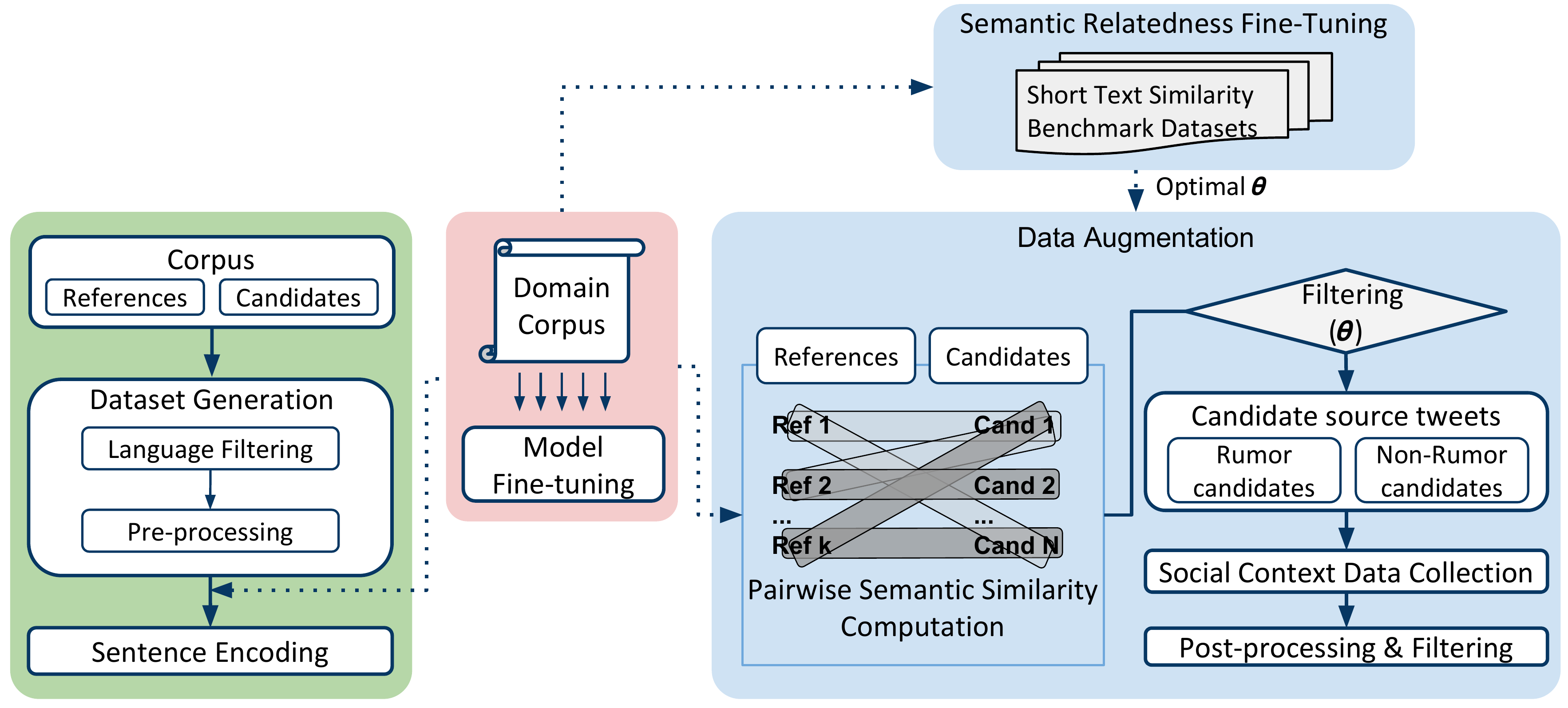}
  \caption{Data augmentation architecture. The leftmost (green) box shows our method for encoding tweet paris using fine-tuned language model. The blue boxes on the right show the key idea: employing fine-tune relatedness thresholds for new source tweets variants identification and rumors dataset generation.}
	\label{methods:overview}
\end{figure*}

An overview of data augmentation method is presented in Figure~\ref{methods:overview}. Input corpus consists of \textit{``References''} and \textit{``Candidates''} sets. \textit{``References''} are limited ground truth source tweets which are exploited to provide higher level supervision for unlabeled candidate tweets (i.e., \textit{``Candidates''}). Candidate tweets refer to any tweets that report an event of interest. Schemes for constructing references vary between data sets. For \textit{PHEME5}, we use annotations in the \textit{PHEME} data. References for the ``Boston marathon bombings" event are generated separately. Detailed reference generation procedure is described in Section~\ref{augref}. 

A deep bidirectional language model (biLM) is first trained with domain-specific corpora in order to learn representations of rumors. We adopt the ELMo biLM model~\citep{Peters2018deep} in our experiments. The leftmost (green) box illustrates data preprocessing and sentence encoding. Given corpora that contain pairs of reference and candidate tweets, we apply language-based filtering and perform linguistic preprocessing. The preprocessing includes lowercasing, the removal of retweet symbols ('rt @'), URLs\footnote{Embedded links can be considered as critical metadata for rumor detection as they may help to identify sources of potential rumors. External information except textual information is not exploited for our data augmentation task.}, and non-alphabetic characters, and tokenization. Tweets with a minimum of 4 tokens are considered to reduce noise~\citep{Ifrim2014Event}. Then, we compute contextual embeddings of tweets using fine-tuned biLM models (see section ~\ref{methods-elmo}). The blue boxes on the right side illustrate our semantic relatedness-based method for rumor variants identification. Cosine distance between embeddings of reference source tweets and those of unlabeled candidate tweets is used as a measurement of semantic similarity. Cosine similarity between vector representation of two sentences is a common metric for measuring semantic similarity \citep{vijayaraghavan2016deepstance}. Two semantically equivalent embeddings have a cosine similarity of 1, and two vectors with no relation have that of 0. To fine-tune relatedness thresholds that determine whether a reference-candidate pair bears strong semantic relation, a standard short-text similarity benchmark dataset (\textit{SemEval-2015 task 1 data}) is used. Two thresholds are learned from the fine-tuning process include a rumor candidate threshold ($\theta_1$) and non-rumor candidate threshold ($\theta_2$). Refer to Section~\ref{thresh-finetune} and ~\ref{exp-augment} for the details of experiment results and our strategy for balancing class distributions. Having optimum thresholds, we compute the pairwise semantic similarity of reference-candidate pairs from the \textit{References} and \textit{Candidates} sets. The next step is to select rumors and non-rumors from candidate tweets based on the optimum relatedness thresholds. In the final step, data collection is performed to retrieve social-temporal context data (typically retweets and replies) for selected candidate tweets. Source tweets without context are filtered out.

\subsection{Reference Generation}\label{augref}
We present how reference data is generated using already available labeled data. For the \textit{PHEME5}, annotated rumor categories in the \textit{PHEME(6392078)} are used. Rumor source tweets are categorized by their topics, and the authors create clean texts for each rumor category. For example, a rumor category for the Sydney siege event,``The Sydney airspace has been closed'', includes several rumor source tweets related to airspace over Sydney. Some examples are as follows: \textbf{(1)} ``\textit{CORRECTION: We reported earlier Sydney air space was shut down. That is not correct. \#SydneySiege}'', and \textbf{(2)} ``\textit{DEVELOPING: Airspace shutdown over Sydney amid chocolate shop hostage situation.}'' We understand that using raw tweets as references may help to capture more various patterns of rumor variations. However, tweets are very noisy and contain a large amount of non-standard spelling. To ensure high quality references and reduce the computation time of pairwise similarity between candidates and references, we use clean rumor categories as rumor references. 

As the ``bostonbombings'' event is not available in the \textit{PHEME(6392078)}, we refer to \textit{CrisisLexT26} as well as the Boston marathon bombings rumor archive created by \url{Snopes.com}. Any rumors investigated by Snopes.com are included in the reference set for ``bostonbombings'' regardless of their veracity. In the \textit{CrisisLexT26}, tweets are categorized by their \textit{informativeness} (related to the crisis and informative, related but not informative, and unrelated), \textit{information type} (affected individuals, affected infrastructure, donations \& volunteers, caution \& advice, emotions, and other useful information), and \textit{information sources} (e.g., eyewitness and media). The original data includes 1000 annotated tweets for the Boston marathon bombings. As the \textit{CrisisLexT26} is not annotated under an annotation scheme for social media rumors, we map its labels to binary labels (i.e., rumors/non-rumors).To this end, tweets with ``related and informative'' \textit{informativeness} label are selected. Next, tweets, the \textit{information type} of which is any of ``affected individuals'', ``infrastructure and utilities'', and ``other useful information'', are chosen. After sampling, 335 annotated tweets remain. We manually inspect and categorize them into rumors and non-rumors according the a rumor tweet annotation scheme proposed by Proter et al. \cite{procter2013reading}. To match the format of references generated using the \textit{PHEME(6392078)}, we generate clean reference sentences from rumor tweets obtained after mapping the \textit{CrisisLexT26} labels and texts available in the Snopes.com's archive. Some examples of references for the `bostonbombings' are as follows:\\
- {\small The third explosion at the JFK library (unknown connection)}\\
- {\small Bombs were pressure cookers and placed in black duffel bags}\\
- {\small Suspect in Boston bombing described as dark skinned male}

\subsection{Data Collection} \label{method-data}
We download source tweets for six selected events in \textit{Twitter events 2012-2016} and \textit{CREDBANK} using an open source tweet collector called \textit{Hydrator \footnote{available via \url{http://github.com/DocNow/hydrator}}}. Table~\ref{tab:event1216} shows the number of tweet ids in the original \textit{Twitter events 2012-2016} data, that of downloaded tweets, that of candidate source tweets which remained after language-based filtering and linguistic preprocessing (see Section~\ref{overview}), and that of references. For \textit{CREDBANK}, \textit{77,954,446} out of \textit{80,277,783} tweets (i.e., 97.1\% of the original data) are downloaded. After deduplication, the train corpus contains \textit{6,157,180} tweets with \textit{146,340,647} tokens and \textit{2,235,075} vocabularies. We collect retweets using a Python library \textit{tweepy} \footnote{available via \url{https://www.tweepy.org/}}. Replies are collected via screen scraping technique implemented using Python libraries \textit{Selenium}~\footnote{available via \url{http://selenium-python.readthedocs.io/}} and \textit{BeautifulSoup}~\footnote{available via \url{http://www.crummy.com/software/BeautifulSoup/bs4/doc/}}.

\begin{table}[t!]
\caption{Statistics of the Twitter events 2012-2016 data.}
\begin{center}
\footnotesize
\begin{tabular}{p{\dimexpr 0.236 \linewidth-2.4\tabcolsep}
				p{\dimexpr 0.21\linewidth-2.8\tabcolsep}
				p{\dimexpr 0.2\linewidth-2.6\tabcolsep}
				p{\dimexpr 0.24\linewidth-2.6\tabcolsep}
				p{\dimexpr 0.18\linewidth-3.0\tabcolsep}}

\bf Event  & \bf \# of tweets (original) &\bf downloaded tweets & \bf after\newline preprocessing & \bf \# of\newline references \\ \hline

germanwings        	& 2,648,983 & 1,726,981 & 702,864& 19\\
sydneysiege      	& 2,157,879 & 1,376,218 & 1,211,295 & 61 \\
fergusonunrest      & 8,782,071 & 5,743,959 & 5,504,692 &41\\
ottawashooting  	& 1,075,864 & 737,136 & 669,734 &51\\
bostonbombings      & 3,430,387 & 1,886,632 & 1,259,857 & 88\\
charlihebdo         & 1,894,0619 & 12,253,734 & 4,276,112 & 60 \\

\end{tabular}
\label{tab:event1216}
\end{center}
\end{table}

\subsection{Rumor-Oriented Embeddings (ELMo)}\label{methods-elmo}

ELMo is adopted to learn effective representation of tweets. ELMo provides deep, contextualized, and character-based word representations by using bidirectional language models (biLMs)~\citep{Peters2018deep}. Previous research shows that fine-tuning NLMs with domain-specific data allows them to learn more meaningful word representations and provides a performance gain~\citep{kim2014convolutional, Peters2018deep}. To fine-tune pretrained ELMo~\footnote{The pretrained model and the Tensorflow training checkpoints are obtained from the Tensorflow implementation of ELMo, available via \url{github.com/allenai/bilm-tf} } for our task, we generate a data set from \textit{CREDBANK}. Sentences in the original corpus are shuffled and split into training and hold-out sets. About 0.02\% of the original data is used as the hold-out set. We also generate a test set using the \textit{PHEME} data containing 6,162 tweets related to 9 events in the hope that it will offer an independent and robust evaluation of our hypothesis (refer to Section~\ref{intro}). As for the \textit{SNAP} corpus, we use ``June'' tweets  as a training set to fine-tune the pretrained ELMo model. We sample 6,000 tweets from ``November'' tweets and use them as a hold-out set. Table~\ref{tab:elmotraining2corpus} shows the number of tweets, tokens and vocabularies in the training and hold-out sets of the \textit{CREDBANK} and \textit{SNAP} after language filtering and deduplication. Following the practice in~\citep{perone2018evaluation}, a linear combination of the states of each LSTM layer and token embeddings is adopted to encode tweets. Since the \textit{CREDBANK} training set is still relatively small for NLMs, we only fine-tune the pre-trained ELMo with one epoch to avoid over-fitting. Table~\ref{tab:perplexitytest2corpus} shows a great improvement in perplexity on both hold-out and test sets with the \textit{CREDBANK} in comparison to the fine-tuned model with the \textit{SNAP}. Reported values are the average of forward and backward perplexity. Once fine-tuned, the biLM weights are fixed and used for computing the sentence representation of tweets in our experiments.

\begin{table}
{\footnotesize\begin{tabular}{p{\dimexpr 0.236 \linewidth-2.4\tabcolsep}
				p{\dimexpr 0.21\linewidth-2.8\tabcolsep}
				p{\dimexpr 0.2\linewidth-2.6\tabcolsep}
				p{\dimexpr 0.24\linewidth-2.6\tabcolsep}
				p{\dimexpr 0.18\linewidth-3.0\tabcolsep}}
\multirow{1}{*}{\bf Corpus} & \bf Item & \bf Train & \bf Hold-out \\ \hline
\multirow{3}{*}{CREDBANK} & tweets & 6,155,948 & 1,232 \\	
    & tokens & 146,313,349 & 27,298 \\
    & vocabs & 2,234,861 & 6,517 \\\hline
\multirow{3}{*}{SNAP} & tweets & 13,928,924 & 6,000\\
    & tokens & 193,192,322 & 99,758 \\
    & vocabs & 11,696,602 & 24585
\end{tabular}}{\caption{Statistics of two corpora for fine-tuning elmo.}
      \label{tab:elmotraining2corpus}}
\end{table}

\begin{table}
\footnotesize
\centering
\begin{tabular}{p{\dimexpr 0.36\linewidth-2\tabcolsep}
				p{\dimexpr 0.14\linewidth-2\tabcolsep}
				p{\dimexpr 0.2\linewidth-1.2\tabcolsep}
				p{\dimexpr 0.18\linewidth-0.2\tabcolsep}}
				
\bf Data & \bf Before tuning & \textbf{After tuning\newline (\scriptsize CREDBANK)} & \bf After tuning\newline  (\scriptsize SNAP)\\\hline
Hold-out (\scriptsize CREDBANK) &883.06 & 18.24 & 389.14\\
\hline
Hold-out (\scriptsize SNAP) & 476.42 & N/A & 68.22 \\
\hline
Test &475.06   & \textbf{32.02} & 311.64
\end{tabular}
\caption{Improvements in perplexity after fine-tuning with two different corpora.}
\label{tab:perplexitytest2corpus}
\end{table}

\section{Experiments}\label{experiments}

\subsection{Semantic Relatedness Fine-Tuning}\label{thresh-finetune}
\begin{figure}\CenterFloatBoxes
\begin{floatrow}
\ffigbox[\FBwidth]{
{\includegraphics[width=.85\linewidth]{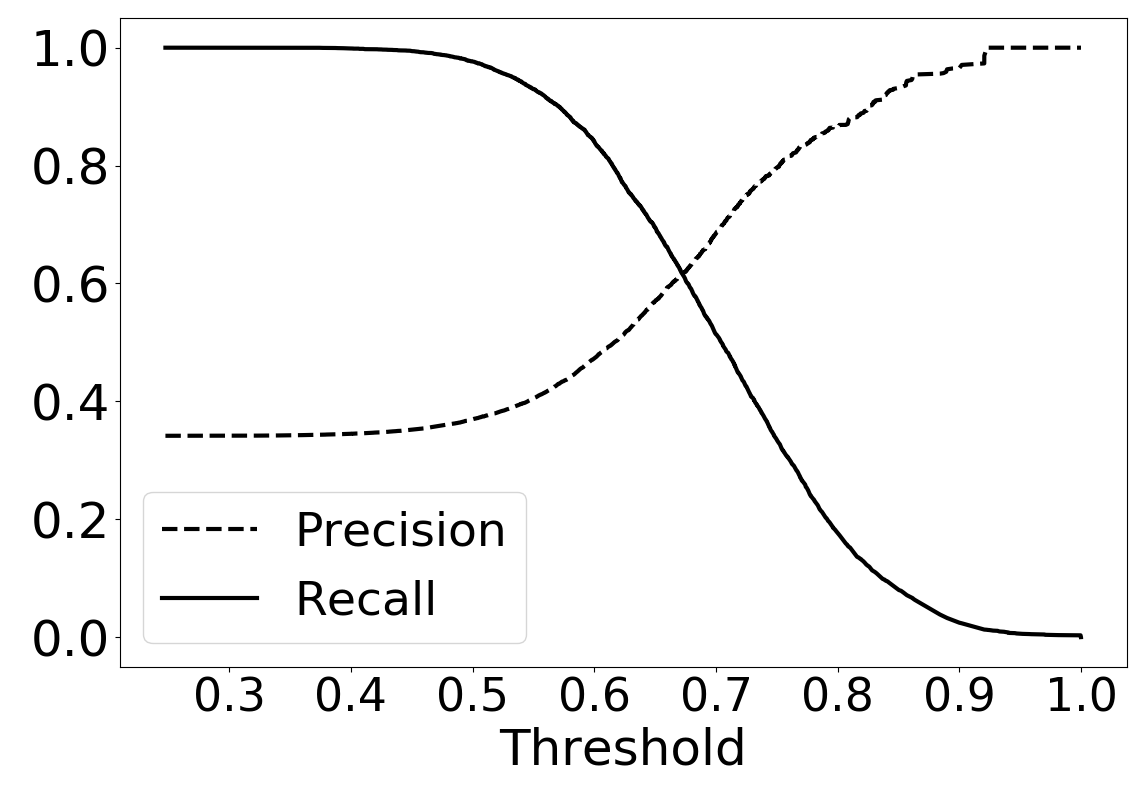} }
}{
	\caption{Precision-recall curve}
}
\killfloatstyle\ttabbox[\Xhsize]{
    \footnotesize
	\begin{tabular}{p{\dimexpr 0.14\linewidth-0.6\tabcolsep}
	p{\dimexpr 0.14\linewidth-0.6\tabcolsep}
	p{\dimexpr 0.14\linewidth-0.6\tabcolsep}
	p{\dimexpr 0.15\linewidth}}
\bf P & \bf F & \bf R & \bf THOLD \\ \hline
0.6088 & 0.6507 &  0.6986 & 0.6526\\
0.7000 & 0.6176 & 0.5526 & 0.6911\\
0.7500 & 0.5907 &  0.4871 & 0.7083\\
0.8502 & 0.4421 &  0.2987 & 0.7602\\
\textbf{0.9003} & 0.2832 &  0.1681 & \textbf{0.8018}\\
\end{tabular}
}
{\caption{Fine-tuning thresholds by precision.}\label{tab:pr-tuning}}
\end{floatrow}
\end{figure}

We are interested in exploring the effect of the distance between embeddings of pairs of reference and candidate tweets on the quality of augmented data which will eventually affect the rumor detection model's capability to predict unseen rumors.
Table~\ref{semeval-results} compares different models for word representation on the SemEval-2015 data. We show the results based on the maximum F-score each model achieved.  Our experimental results show the effectiveness of our CREDBANK fine-tuned ELMo over pre-trained model ("Original (5.5B)") and SoA word embedding models. We applied different models to normalized texts. Normalization methods we used in the experiments include removing English stopwords and punctuations, and lemmatization using `WordNetLemmatizer' in a Python library NLTK~\footnote{available via \url{https://www.nltk.org}}. As shown in Table~\ref{semeval-results}, text normalization actually degenerates the performance of the ELMo in terms of F-score, while it improves the performance of the other word embeddings. In fact, state-of-the-art NLMs like ELMo do not need much text normalization. A pre-trained ELMo model only needs tokenization. As for the output of ELMo models, using the average of representations from all layers outperforms using only the top layer representation. This finding is consistent with results presented in Perone et al.'s work~\citep{perone2018evaluation}. To ensure higher quality (i.e., less false positives in a selected sample),   relatedness thresholds are fine-tuned based on precision achieved by the best-performing model. Table~\ref{tab:pr-tuning} shows a part of fine-tuning results. We should choose a threshold which can achieve a reasonably high precision and sample an adequate number of tweets.

\begin{table}[t!]
\caption{Comparison of the paraphrase identification performance of different models for sentence representation.}
\label{semeval-results}
\begin{center}
\footnotesize
\begin{tabular}{p{\dimexpr 0.45\linewidth-2.0\tabcolsep}
				p{\dimexpr 0.13\linewidth-2.0\tabcolsep}
				p{\dimexpr 0.11\linewidth-2.0\tabcolsep}
				p{\dimexpr 0.11\linewidth-2.0\tabcolsep}
				p{\dimexpr 0.13\linewidth}}

\bf Model  & \bf F & \bf P & \bf R &\bf Threshold
\\ \hline
ELMo+CREDBANK (average)         & \bf{0.6507} & \bf{0.6088} & 0.6986 & 0.6526\\
ELMo+CREDBANK (top)         	& 0.6270 & 0.5660 & 0.7027 & 0.6470\\
ELMo Original 5.5B (average)    & 0.6281 & 0.5872 & 0.6752 & 0.6305\\
ELMo Original 5.5B (top)        & 0.6047 & 0.5554 & 0.6635 & 0.6875\\
GloVe (twitter.27B.200d)        & 0.5079 & 0.3417 & \bf{0.9890} & 0.5017\\
Word2Vec (Google News)          & 0.4223 & 0.4796 & 0.3772 & 0.5003\\
ELMo Original 5.5B (top)$^{\mathrm{*}}$     	& 0.5868 & 0.5112 & 0.6887 & 0.6752\\
GloVe (twitter.27B.200d)$^{\mathrm{*}}$       & 0.5117 & 0.3565 & 0.9062 & 0.5070\\
Word2Vec (Google News)$^{\mathrm{*}}$ 		& 0.4715 & 0.4473 & 0.4985 & 0.5000\\
\hline
\multicolumn{5}{l}{$^{\mathrm{*}}$Models are applied to normalized tweets.}

\end{tabular}
\end{center}
\end{table}

\subsection{Data Augmentation}\label{exp-augment}
We follow our data augmentation procedure described in Section~\ref{methods}. 
After pairwise similarity computation on all references and candidates, we apply relatedness thresholds to the results for selecting rumor and non-rumor source tweets from a pool of candidates. For sampling rumor sources, we use $\theta_1=\textit{0.8018}$, which achieves a precision of 0.9 in the benchmark task illustrated above. If a semantic similarity score between a candidate and one or more references is greater than or equal to $\theta_1$, the candidate is included in a rumor source collection. If a candidate is identified as a rumor for any of rumor references, this candidate is included in a rumor. For non-rumor sources, we assume that low semantic relatedness to rumor references indicate the high likelihood of being a non-rumor. The minimum semantic similarity score for positive paraphrase pairs in the SemEval-2015 task is 0.248. We set a threshold $(\theta_2)$ for sampling non-rumor samples to 0.266, which is the second smallest semantic similarity score for the SemEval-2015 task and achieves the same precision, recall, and F-measure as the minimum score 0.248. If a semantic similarity between a candidate and every rumor reference is less than $(\theta_2)$, the candidate is included in a non-rumor source collection. Data augmentation results after applying thresholds show high class imbalance for all event except the `germanwings'. To overcome this problem, random sampling is applied to the non-rumor source collection. Specifically, we randomly sample $(3*(\texttt{number of augmented rumor sources}))$ non-rumors from the collection. Given augmented and initially balanced rumor and non-rumor source tweets, replies for each source tweet are collected (see Section~\ref{method-data}) and source tweets without replies are removed from the augmented data. We observe a considerable reduction in augmented data size because a large number of source tweets do not have replies. Next, we apply sampling again. $(2*(\texttt{number of rumor source tweets}))$ non-rumor source tweets are randomly sampled to balance class distributions in each event data set. In order to keep source tweets which are rich in conversational threads, we include all source tweets that have more than 10 replies. The remainder is randomly chosen. Finally, augmented rumor and non-rumor source tweets with replies are merged with the \textit{PHEME5}.

\subsection{Rumor Detection}\label{rumour_detection}
We conduct rumor detection experiments using the original \textit{PHEME5} and two augmented data sets: \textit{PHEME5+Aug} and \textit{PHEME5+Aug+boston}. \textit{PHEME5+Aug} is augmented data for the five events in the \textit{PHEME5}. \textit{PHEME5+Aug+boston} is \textit{PHEME5+Aug} combined with the ``bostonbombings''. We employ Kochkina et al.~\citep{kochkina2018all}'s method as a SoA baseline model of rumor detection with three modifications \footnote{We make our source code and experimental datasets data publicly available via \url{https://github.com/soojihan/Multitask4Veracity}}. In their model, source tweets and replies are represented as 300-dimensional Word2Vec word embeddings pre-trained on the Google News data set~\footnote{\url{https://code.google.com/archive/p/word2vec/}}. For the sake of simplicity, we modify the implementation of MTL2 Veracity+Detection~\footnote{available via \url{http://github.com/kochkinaelena/Multitask4Veracity}} for rumor detection only. Another modification we made is data input. In the original models, a conversation consists of a source tweet and replies to it and conversations are decomposed into branches. In our experiments, we are unable to obtain the conversation structure and decompose it into several branches with our augmented datasets. For example, if tweet B is a reply of a source tweet A and tweet C is a reply of B, Twitter objects represent that C is a reply of A. To overcome this constraints but still take contexts into consideration, we consider the entire conversation of a source tweets as a single branch. We construct input by using source tweet and the top (i.e., most recent) 24 replies of each source tweet in this task. The original models require input with shape: (the number of branches in each event dataset, the maximum length of branches, 300). Therefore, the modified models require input with shape: (the number of source tweets in each event data, 25, 300). Finally, training and hold-out sets are generated independently from test sets (see details as follows). In the original implementation, validation sets used for hyperparameter optimization during the training are the same as test set, which results in biased evaluation. In addition, we implement the grid search with the parameter space defined in the original work \citep{kochkina2018all}, which runs with 30 trials.

In order to evaluate the performance and ability of the model using the augmented data in a realistic scenario, we adopt leave-one-out cross-validation (LOOCV) method. Simply speaking, one event is used as a test set and the remaining events are used as a training set on each iteration. This allows us to verify whether the augmented data can help to generalize rumor detection model to unseen rumors. For the \textit{PHEME5}, four out of five original PHEME5 events are shuffled and split into training and hold-out sets. The remaining one event is used as a test set for evaluation. Thus, a 5-fold CV is applied. The \textit{PHEME5+Aug} and \textit{PHEME5+Aug+boston} are also evaluated in a 5-fold CV setting as the \textit{PHEME5}. The only difference is that training and hold-out sets are generated using the augmented data. In other words, test sets generated from the \textit{PHEME5} is used for all three settings. Class distribution of training, validation, and test sets are equally balanced. This helps to evaluate the contribution of data augmentation on rumor detection by mitigating class imbalance.

\section{Results and Discussion}\label{results}
\subsection{Data Augmentation}

\begin{table*}[h!]
	\caption{Number of rumor and non-rumor source tweets and replies in the augmented data.}
	\label{eval:aug-results}
	\footnotesize
	\centering
	\begin{tabular}{
			p{\dimexpr 0.12\linewidth-1.2\tabcolsep}
			|p{\dimexpr 0.06\linewidth-1.2\tabcolsep} 
			p{\dimexpr 0.07\linewidth-1.2\tabcolsep}
			p{\dimexpr 0.07\linewidth-1.2\tabcolsep}
			p{\dimexpr 0.07\linewidth-1.2\tabcolsep}|
			p{\dimexpr 0.1\linewidth-1.2\tabcolsep}
			p{\dimexpr 0.12\linewidth-1.2\tabcolsep}
			p{\dimexpr 0.11\linewidth-1.2\tabcolsep} 
			p{\dimexpr 0.13\linewidth-1.2\tabcolsep} }
		
		& \multicolumn{4}{c|}{\bf Augmented data} & \multicolumn{4}{c}{\bf After balancing and merging} \\ 
		\hline
		& \multicolumn{2}{c}{\textbf{Rumor}} & \multicolumn{2}{c|}{\textbf{Non-rumor}} & \multicolumn{2}{c}{\textbf{Rumor}} & \multicolumn{2}{c}{\textbf{Non-rumor}} \\ 
		
		\textbf{Event} & \textbf{source} & \textbf{threads} & \textbf{source} & \textbf{threads} & \textbf{source} & \textbf{threads} & \textbf{source} & \textbf{threads} \\ \hline 
		germanwings & 272 & 1,028 & 373 & 1,099 & 502 (238) & 3,231 (2,256) & 604 (231) & 2,863 (1,764)\\
		sydneysiege  & 1,289 & 4,632 & 3,955 & 14,673 & 1,764 (522)& 12,330 (8,155) & 3,530 (699) & 27,797 (14,621) \\
		ottawashooting & 625 & 3,335 & 3,607 & 18,340 & 1,047 (470)  & 8,860 (5,966) & 2,072 (420) & 20,933 (5,428) \\
		ferguson & 475 & 2,222 & 2,934 & 11,168 & 737 (284)  & 8,184 (6,196) & 1,476 (859) & 24,639 (16,837)\\
		charliehebdo & 802 & 3,565 & 4,437 & 22,969 & 1,225 (458) & 10,152 (6,888) & 2,450 (1,621) & 45,765 (29,302) \\ 
		bostonbombings & 429 & 2,084 & 3,231 &  31,290 & 392 (N/A) & 2,084 (N/A) & 784 (N/A)& 24,536 (N/A)\\
		\hline
		\multicolumn{1}{r|}{\textbf{Total}} & 3,892 & 16,866 & 18,537 & 100,439 & 5,667 (1,972) & 44,805 (29,461) & 10,916 (3,830) & 146,533 (67,952) 
	\end{tabular}
\end{table*}

\begin{table}[h!]
\caption{Data statistics after temporal filtering and deduplication.}
\label{eval:aug-temp}
\footnotesize
\centering
\begin{tabular}{
				p{\dimexpr 0.22\linewidth-1.6\tabcolsep}
				p{\dimexpr 0.14\linewidth-1\tabcolsep}
			    p{\dimexpr 0.08\linewidth-1.3\tabcolsep} 
				p{\dimexpr 0.09\linewidth-1.3\tabcolsep}
				p{\dimexpr 0.06\linewidth-1.1\tabcolsep}
				p{\dimexpr 0.08\linewidth-1.3\tabcolsep}
				p{\dimexpr 0.09\linewidth-1.3\tabcolsep}
				p{\dimexpr 0.06\linewidth-0.3\tabcolsep}}

 & & \multicolumn{3}{c}{\textbf{Rumor}} & \multicolumn{3}{c}{\textbf{Non-rumor}} \\ \hline

\textbf{Event} & \textbf{Date} &\textbf{source} & \textbf{thread} & \textbf{Mdn}& \textbf{source} & \textbf{thread} & \textbf{Mdn}\\ \hline 
germanwings & 30/24/2015 &375 & 2,801 & 4(7) & 402 & 2,202 & 3(5)\\
sydneysiege  & 12/14/2014 &1,134 & 10,271& 4(15) & 2,262 & 19,547 & 3(18) \\
ottawashooting & 10/22/2014 &713 & 7,117 & 6(11) & 1,420 & 10,522 & 4(10) \\
ferguson & 08/09/2014 &471 & 7,103 & 8(17) & 949 & 19,545 & 15(14)\\
charliehebdo & 01/07/2015 &812 & 8,356 &5(12)& 1,673 & 34,435 & 16(15) \\ 
bostonbombings & 04/15/2013 &323 & 1,973 &2(\textendash)& 645 & 21,871 & 20(\textendash) \\
\hline
\multicolumn{1}{r}{\textbf{Total}} & & 3,828 & 37,621 & \textendash & 7,351 & 108,122 & \textendash
\end{tabular}
\end{table}

As illustrated in Section~\ref{experiments}, we augment rumor and non-rumor source tweets for the six selected events in the Twitter events 2012-2016 data. Then, the augmented tweets for the PHEME5 events are merged with the original PHEME5. Table~\ref{eval:aug-results}\footnote{Statistics of enriched retweets are omitted in this paper due to the scope constraints of this paper. See augmented rumor corpus website for details.} shows the number of source tweets and replies obtained via our data augmentation method and those after balancing augmented data and merging the balanced data with the original PHEME5.  The number of conversational threads in original PHEME5 is provided in parentheses for comparison. Overall, the number of source tweets for rumors and non-rumors increased by 187\% and 185\%, respectively. There are 52\% and 110\% increases in the number of replies for rumor sources and that for non-rumor sources, respectively. The standard deviation of imbalance ratios of non-rumor sources to rumor source improved from 1.24\% to 0.35\%, respectively. In particular, significant class imbalances in two largest events--``fergusonunrest'' and ``charliehebdo''--have become moderate as a result of data augmentation.

A manual inspection of sampled source tweets shows that augmented data contains tweets identical to references and several variations of references. It is worth noting that our data augmentation with weak supervision can even capture rumors which are related but not technically identical to reference tweets. Some examples of rumor tweets in our augmented data are as follows: \\
\textbf{(1)} \textit{\textbf{A 20-year-old student} is among the hostages at the kosher shop in Paris \url{http://t.co/orBfH8MK1J}}: This tweet is almost identical to a reference tweet, ``\textbf{A baby} is among the hostages in the Kosher market'', for the Charlie Hebdo attack, except for subjects of sentences. The semantic similarity score between two sentences is 0.8123. \\
\textbf{(2)} \textit{Uber Promises \textbf{Free Rides} in Sydney after Surge Pricing Kicks in During Hostage Crisis \url{http://t.co/7NAO9HSxEA}}: This tweet is a variation of a reference tweet, ``Uber introduced \textbf{surge pricing} in downtown Sydney during hostage crisis.''. Two sentences report contradictory sub-events related to a taxi booking company called Uber, but their semantic similarity score is 0.8238.

Using raw annotated tweets as references rather than refined categories of rumors may help to retrieve more positive examples. In the original PHEME, for example, a tweet, ``Ray Hadley says he spoke with hostage, and could hear the gunman in the background barking orders and demanding to go live on air'', is annotated as a rumor, ``The gunman and/or hostages have made contact with Sydney media outlet(s) (radio station, etc.)''. Without a background knowledge that Ray Hadley is an Australian radio broadcaster, data augmentation methods based on semantic relatedness fail to identify such rumors.


\subsection{Rumor Detection}

Since we are concerned with the usefulness of data augmentation for early rumor detection, temporal filtering is applied on every individual event data set in our experiments. Specifically, we only retain source tweets and social context within 7 days after each event occurs. Table \ref{eval:aug-temp} shows the details of augmented rumor corpus and the median (Mdn) number of conversation threads after temporal filtering. The median values of original PHEME data are given in parentheses for comparison. 

Table~\ref{eval:detection-results} shows the overall performance of rumor detection with three different data sets. The values of four evaluation metrics are the mean scores of all LOOCV iterations. The overall results show that data augmentation helps to boost performance on rumor detection task in terms of F-score (F), precision (P), recall (R) and accuracy (Acc.). The model performance in terms of F-score increases by 9\% and 12\% with \textit{PHEME5+Aug} and \textit{PHEME5+Aug+boston}, respectively. Table~\ref{eval:loocv} shows the details of LOOCV results described in Section~\ref{rumour_detection}. ``Event'' column in Table~\ref{eval:loocv} shows 5 different events used as a test set on each iteration of LOOCV. It is worth noting that the Ferguson unrest was the most difficult event in the \textit{PHEME5} for a rumor detection model as it has a unique class distribution distinguished from all other events~\citep{kochkina2018all}. With augmented dataset \textit{PHEME5+Aug+boston}, the F-measure on this event increases by 36.7\%.

\begin{table}[ht!]
\caption{Overall rumor detection performance of three data sets.}
\label{eval:detection-results}
\footnotesize
\centering
\begin{tabular}{p{\dimexpr 0.43\linewidth-2.5\tabcolsep}
				p{\dimexpr 0.15\linewidth-2.5\tabcolsep}
				p{\dimexpr 0.15\linewidth-2.5\tabcolsep} 
				p{\dimexpr 0.15\linewidth-2.5\tabcolsep} 
				p{\dimexpr 0.15\linewidth-2.5\tabcolsep} }
\bf Data &  \bf F & \bf P & \bf R & \bf Acc. \\ \hline 
\textbf{PHEME5} & 0.535 & 0.650 & 0.484 & 0.622 \\
\textbf{PHEME5+Aug} &  0.625 &  0.688 & 0.585 & 0.664 \\
\textbf{PHEME5+Aug+boston} &  \textbf{0.656} & \textbf{0.716} & \textbf{0.614} & \textbf{0.685} \\ 
\end{tabular}
\end{table}

\begin{table}[h!]
\caption{LOOCV results for the PHEME5 and augmented data sets.}
\label{eval:loocv}
\footnotesize
\centering
\begin{tabular}{p{\dimexpr 0.22\linewidth-1.3\tabcolsep}
				p{\dimexpr 0.3\linewidth-1.3\tabcolsep} 
				p{\dimexpr 0.12\linewidth-2.5\tabcolsep}
				p{\dimexpr 0.12\linewidth-2.5\tabcolsep} 
				p{\dimexpr 0.12\linewidth-2.5\tabcolsep} 
				p{\dimexpr 0.12\linewidth-2.5\tabcolsep}}
\bf Event & \bf Data & \bf F & \bf P & \bf R & \bf Acc. \\ \hline 
\multirow{3}{*}{\bf charliehebdo} & PHEME5 & 0.758 & 0.714 &   0.808 & 0.742 \\
								& PHEME5+Aug & 0.742 &  \textbf{0.734} & 0.749 & 0.739  \\
								& PHEME5+Aug+boston &  \textbf{0.767} & 0.723 & \textbf{0.817} &  \textbf{0.752} \\ \hline

\multirow{3}{*}{\bf sydneysiege} & PHEME5 & 0.583 & 0.714  & 0.492 & 0.648 \\
								& PHEME5+Aug &  \textbf{0.695} &  \textbf{0.755} &  \textbf{0.644} &  \textbf{0.717} \\
								& PHEME5+Aug+boston & 0.632 & 0.759  & 0.542 & 0.685 \\ \hline

\multirow{3}{*}{\bf fergusonunrest} & PHEME5 & 0.242 & 0.550 & 0.155 & 0.514  \\
								& PHEME5+Aug & 0.416  & 0.618  & 0.313  & 0.560 \\
								& PHEME5+Aug+boston &  \textbf{0.609} &  \textbf{0.707} &  \textbf{0.535} &  \textbf{0.657} \\ \hline

\multirow{3}{*}{\bf ottawashooting} & PHEME5 & 0.516 & 0.653 & 0.426 & 0.600 \\
								& PHEME5+Aug & 0.671 & 0.680 &  \textbf{0.662} & 0.675 \\
								& PHEME5+Aug+boston &  \textbf{0.697} &  \textbf{0.739} & 0.660 &  \textbf{0.713} \\ \hline

\multirow{3}{*}{\bf germanwings} & PHEME5 &  0.577 & 0.619 & 0.541 & 0.604 \\
								& PHEME5+Aug &  \textbf{0.601} &  \textbf{0.652} &  \textbf{0.558} &  \textbf{0.630} \\
								& PHEME5+Aug+boston & 0.575 & 0.650 & 0.515 & 0.619
\end{tabular}
\end{table}

\section{Conclusions and Future Work}\label{conclusions}
In this paper, we have proposed a new paradigm of data augmentation for effectively enlarging existing rumor data sets using publicly available large-scale unlabeled data associated with real-world events. Semantic relatedness is exploited to annotate unlabeled data with weak supervision based on limited labeled rumor source tweets. Our experiments show the potential efficiency and effectiveness of semantically augmented data for combating the scarcity of labeled data and class imbalance of existing publicly available rumor data sets. Our augmented data is highly realistic, can potentially increase the diversity of existing labeled data, and can improve its quality. Preliminary results achieved using a SoA DNN model indicate that augmented training data is helpful to train DNNs by preventing them from overfitting, and consequently improves model generalization. We release our augmented data in the hope that it will be useful for further research in the field of rumor detection and general studies of rumor propagation on social networks. In the future, we plan to extend our method to other events and training tasks in order to build more comprehensive data for rumor detection. A more extensive evaluation will be conducted to examine the effectiveness of augmented data in handling over-fitting and its usefulness in facilitating deeper NNs for rumor detection. Further research will also look into more advanced techniques for rumor variation identification and conduct more comprehensive evaluation with other NLMs such as GPT and BERT. In addition, it is arguable that different types of rumor events may expose different propagation patterns. We will look into whether data augmentation creates a bias towards detecting the same sort of rumors. Increasing diversity and reducing bias in training data will be a future direction of our work.

\bibliography{main}
\bibliographystyle{IEEEtran}

\end{document}